\pdfoutput=1

\documentclass[11pt]{article}

\usepackage[final]{emnlp2021}

\usepackage{times}
\usepackage{latexsym}

\usepackage[T1]{fontenc}

\usepackage[utf8]{inputenc}

\usepackage{microtype}

\usepackage{microtype}
\usepackage{graphicx}
\usepackage{booktabs} 
\usepackage{makecell}
\usepackage{graphicx}
\usepackage{caption} 
\usepackage{courier}
\usepackage{multirow}
\usepackage{color}
\usepackage{subcaption}
\usepackage{color}
\usepackage{amssymb}
\usepackage{gensymb}
\usepackage{amsfonts}       
\usepackage{enumitem}
\usepackage{multirow}
\usepackage{algorithm}
\usepackage{algorithmic}
\usepackage{amsmath,nccmath}
\usepackage{hyperref}

\title{Knowledge-Aware Meta-learning for Low-Resource Text Classification}

\author{Huaxiu Yao$^{1\dag}$, Yingxin Wu$^{2\ddag}$, Maruan Al-Shedivat$^{4	\S}$, Eric P. Xing$^{3,4\S}$ \\
  $^1$Stanford University; $^2$University of Science and Technology of China\\
  $^3$Mohamed bin Zayed University of Artificial Intelligence,
  $^4$Carnegie Mellon University
\\
  \texttt{$^\dag$huaxiu@cs.stanford.edu, $^\ddag$wuyxinsh@gmail.com}\\
  \texttt{$^\S$\{alshedivat, epxing\}@cs.cmu.edu}}

\date{}

\begin{document}
\maketitle

\begin{abstract}
Meta-learning has achieved great success in leveraging the historical learned knowledge to facilitate the learning process of the new task. However, merely learning the knowledge from the historical tasks, adopted by current meta-learning algorithms, may not generalize well to testing tasks when they are not well-supported by training tasks. This paper studies a low-resource text classification problem and bridges the gap between meta-training and meta-testing tasks by leveraging the external knowledge bases. Specifically, we propose KGML to introduce additional representation for each sentence learned from the extracted sentence-specific knowledge graph. The extensive experiments on three datasets demonstrate the effectiveness of KGML under both supervised adaptation and unsupervised adaptation settings.
\end{abstract}
\section{Introduction}
Learning-to-learn (or meta-learning) \citep{bengio1990learning, schmidhuber1992learning, hochreiter2001learning, vinyals2016matching, finn2017model} has recently emerged as a successful technique for training models on large collections of low-resource tasks.
In the natural language domain, it has been used to improve machine translation \citep{gu2018meta}, semantic parsing \citep{sun2020neural}, text classification \citep{bao2019few, geng2020dynamic, geng2019induction,li2020learn}, sequence labelling~\citep{li2021meta}, text generation~\citep{guo2020inferential}, knowledge graph reasoning~\citep{wang2019meta}, among many other applications in low-resource settings.

Meta-learning has been shown to dominate self-supervised pretraining techniques such as masked language modeling \citep{devlin2018bert} when the training tasks are representative enough of the tasks encountered at test time~\citep{bansal2019learning, bansal2020self}.
However, in practice, it requires access to a very large number of training tasks \citep{alshedivat2021data} and, especially in the natural language domain, mitigating discrepancy between training and test tasks becomes non-trivial due to new concepts or entities that can be present at test time only.

\begin{figure}[t]
    \centering
    \includegraphics[width=\columnwidth]{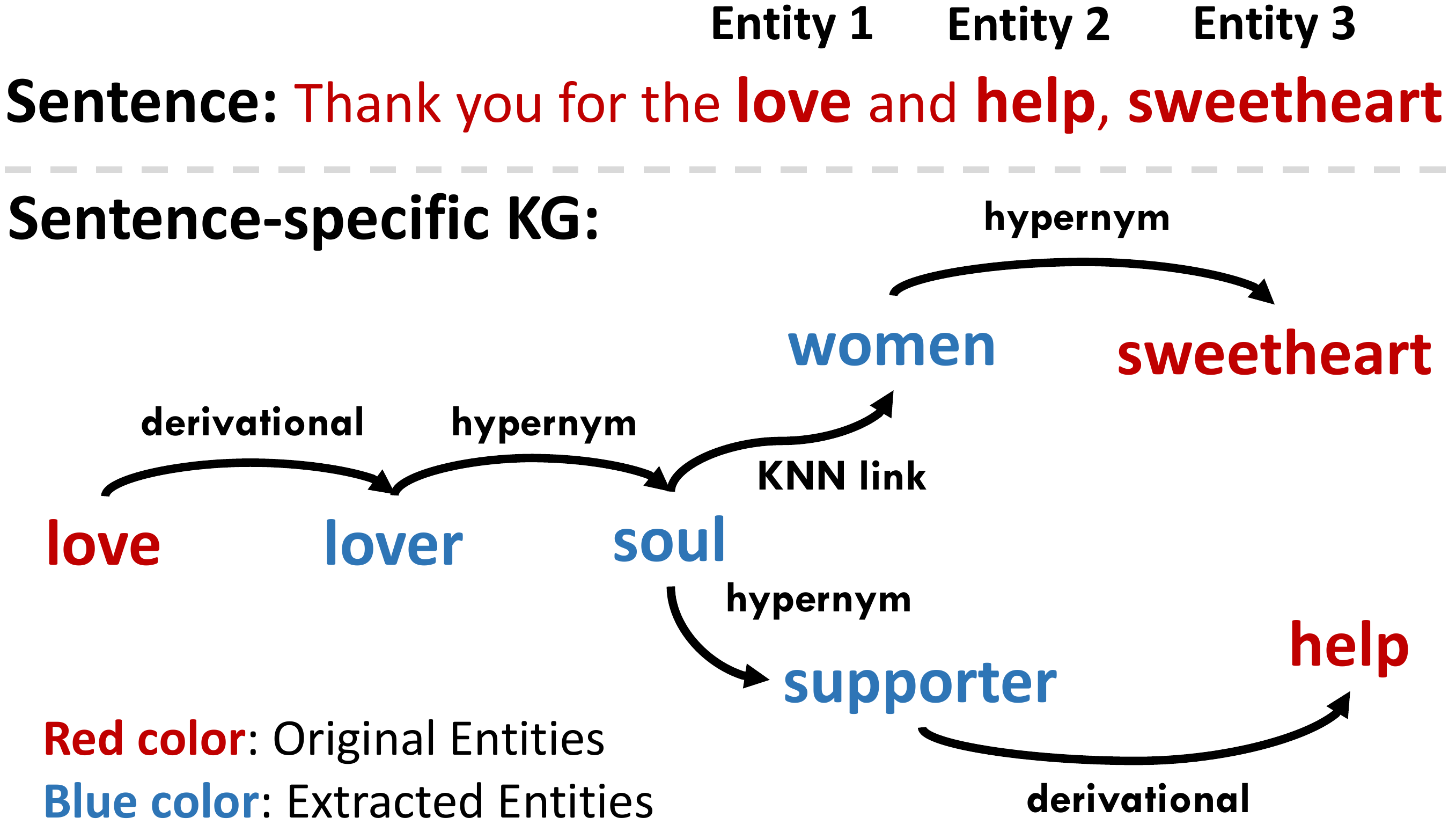}
    \caption{Illustration of extracting a sentence-specific KG from a shared KB.}
    \label{fig:sentence-to-kg}
    \vspace{-1.6em}
\end{figure}

In this paper, we propose to leverage external knowledge bases (KBs) in order to bridge the gap between the training and test tasks and enable more efficient meta-learning for low-resource text classification.
Our key idea is based on computing additional representations for each sentence by constructing and embedding sentence-specific knowledge graphs (KGs) of entities extracted from a knowledge base shared across all tasks (e.g., Fig.~\ref{fig:sentence-to-kg}).
These representations are computed using a graph neural network (GNN) which is meta-trained end-to-end jointly with the text classification model.
Our approach is compatible with both supervised and unsupervised adaptation of predictive models.

\paragraph{Related work.}
In modern meta-learning, there are two broad categories of methods:
(i) gradient-based \citep{finn2017model, nichol2018reptile, li2017meta, zhang2020adaptive,zintgraf2019fast,lee2018gradient,yao2019hierarchically,yao2020automated} and (ii) metric-based \citep{vinyals2016matching, snell2017prototypical, yang2018learning, yoon2019tapnet, liu2019few, sung2018learning}.
The first category of methods represents the ``meta-knowledge'' (i.e., a transferable knowledge shared across all tasks) in the form of an initialization of the base predictive model.
Methods in the second category represent meta-knowledge in the form of a shared embedding function that allows to construct accurate non-parametric predictors for each task from just a few examples. Both classes of methods have been applied to NLP tasks \citep[e.g.,][]{han2018fewrel, bansal2019learning, gao2019fewrel}, however, methods that can systematically leverage external knowledge sources typically available in many practical settings are only starting to emerge and focusing on limited applicable scopes~(e.g., \citep{qu2020few,seo2020physics}).

\paragraph{Contributions.}
\begin{enumerate}[itemsep=-2pt,leftmargin=15pt,topsep=2pt]
    \item We investigate a new meta-learning setting where few-shot tasks are complemented with access to a shared knowledge base (KB).
    \item We develop a new method (KGML) that can leverage an external KB and bridge the gap between the training and test tasks.
    \item Our empirical study on three text classification datasets (Amazon Reviews, Huffpost, Twitter) demonstrates the effectiveness of our approach.
\end{enumerate}
\begin{figure*}[t]
\centering
	\includegraphics[width=\textwidth]{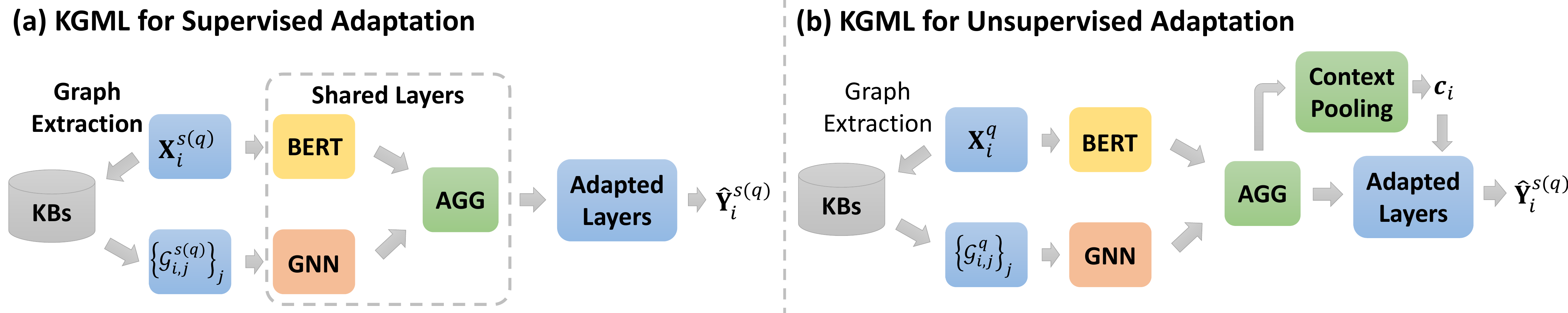}
	\caption{KGML framework on (a) supervised and (b) unsupervised adaptation settings. AGG represents the aggregator $\mathrm{AGG}_{kf}$ for knowledge fusion.}
	\label{fig:framework}
	\vspace{-1em}
\end{figure*}
\section{Preliminaries}
\newcommand{\task}{\mathcal{T}}
\newcommand{\taski}{\task_i}
\newcommand{\testtask}{\task_t}
\newcommand{\data}{\mathcal{D}}
\newcommand{\datais}{\mathcal{D}_i^s}
\newcommand{\dataiq}{\mathcal{D}_i^q}
\newcommand{\featis}{\mathbf{X}_i^s}
\newcommand{\featiq}{\mathbf{X}_i^q}
\newcommand{\labelis}{\mathbf{Y}_i^s}
\newcommand{\labeliq}{\mathbf{Y}_i^q}
\newcommand{\protor}{\mathbf{c}_r}

\newcommand{\datas}{\mathcal{D}^s}
\newcommand{\dataq}{\mathcal{D}^q}
\newcommand{\feats}{\mathbf{X}^s}
\newcommand{\featq}{\mathbf{X}^q}
\newcommand{\labels}{\mathbf{Y}^s}
\newcommand{\labelq}{\mathbf{Y}^q}
\newcommand{\hiddenis}{\mathbf{H}_i^s}
\newcommand{\hiddeniq}{\mathbf{H}_i^q}
\newcommand{\hiddens}{\mathbf{H}^s}
\newcommand{\hiddenq}{\mathbf{H}^q}
\newcommand{\loss}{\mathcal{L}}

\newcommand{\ie}{\emph{i.e., }}
\newcommand{\eg}{\emph{e.g., }}
\newcommand{\etal}{\emph{et al.}}
\newcommand{\st}{\emph{s.t. }}
\newcommand{\etc}{\emph{etc.}}
\newcommand{\wrt}{\emph{w.r.t. }}
\newcommand{\cf}{\emph{cf. }}
\newcommand{\aka}{\emph{aka. }}

We consider the standard meta-learning setting, where given a set of training tasks $\task_1, \dots, \task_n$, we would like to learn a good parameter initialization $\theta_\star$ for a predictive model $f_\theta$ such that it can be quickly adapted to new tasks given only a limited amount of data (i.e., few-shot regime).
Each task $\taski$ has a support set of labeled or unlabeled sentences $\datas_i=\{\feats_i, \labels_i\}=\{(\mathbf{x}_{i,j}^s, \mathbf{y}_{i,j}^s)\}_{j=1}^{N^s}$ and a query set, $\dataq_i=\{\featq_i, \labelq_i\}=\{(\mathbf{x}_{i,j}^q, \mathbf{y}_{i,j}^q)\}_{j=1}^{N^q}$ of labeled sentences.

In our text classification setup, we assume that parameters $\theta$ are split into two subsets: (1) BERT \citep{devlin2018bert} parameters $\theta^{B}$ shared across tasks and (2) task-specific parameters $\theta^c$ that are adapted for each task.
Below, we discuss two adaptation strategies: \emph{supervised} and \emph{unsupervised}.

\subsection{Supervised adaptation}
Under supervised adaptation scenario, we incorporate knowledge with both gradient-based meta-learning and metric-based meta-learning, which are detailed as:

\noindent \textbf{Gradient-based meta-learning.} Following \citet{finn2017model}, the task-specific parameters $\theta^c_i$ for each task $\task_i$ can be adapted by finetuning them on the support set: $\theta^c_i = \theta^c - \alpha\nabla_{\theta^c} \loss(f_{\theta^c,\theta^B}; \datais)$,
where $\loss$ is the cross-entropy loss.
Then, using the query set $\dataiq$, we can evaluate the post-finetuning model and optimize the model initialization as follows:
\begin{equation}
    \theta^{c}_\star, \theta^{B}_\star \leftarrow \arg\min_{\theta^c, \theta^B}\frac{1}{n}\sum_{i}\loss(f_{\theta^c_i,\theta^B};\dataiq)
\end{equation}
At evaluation time, the initialization parameters $\theta_\star$ are adapted to test tasks $\task_t$ by finetuning on the corresponding support sets $\datas_t$.

\noindent \textbf{Metric-based Meta-learning.} Following~\cite{snell2017prototypical} Prototypical Network (ProtoNet), the task-specific parameter $\theta_i^c$ is formulated as a lazy classifier, which is built upon the prototypes $\mathbf{c}_i^k=\frac{1}{|\mathcal{D}_{i,k}^s|}\sum_{j}f_{\theta^B}(\mathbf{x}_{i,j;k}^s)$. Here, $\mathcal{D}_{i,k}^s$ represents the subset of support sentences belonging to class $k$. Then, for each sentence in the query set, the probability of assigning it to class $k$ is calculated as:
\begin{equation}
  p(\mathbf{y}_{i,j}^q=k|\mathbf{x}_{i,j}^q)=\frac{\exp(-d(f_{\theta^B}(\mathbf{x}_{i,j}^{q}),\mathbf{c}_i^k))}{\sum_{k'}\exp(-d(f_{\theta^B}(\mathbf{x}_{i,j}^{q}),\mathbf{c}_i^{k'}))},
\end{equation}
where $d$ is defined as a distance measure. During the meta-training phase, ProtoNet learns a well-generalized embedding function $\theta^B_{\star}$. Then, the meta-learned $\theta^B_{\star}$ is applied to the meta-testing task, where each query sentence is assigned to the nearest class with the highest probability  (i.e., $\hat{\mathbf{y}}_{t,j}^q=\arg\max_{r} p(\mathbf{y}_{t,j}^q=r|\mathbf{x}_{t,j}^q)$). 

\subsection{Unsupervised adaptation}
When labeled supports sets $\datais$ are not available, we follow \citet{zhang2020adaptive} and use ARM-CML.
For each task $\taski$, we use the shared BERT encoder to compute a representation of each query sentence $\mathbf{x}_{i,j}^q$, which returns an embedding vector, denoted $f_{\theta^{B}}(\mathbf{x}_{i,j}^q)$.
Then, we compute the overall representation of the task by averaging these embedding vectors, $\mathbf{c}_i=\frac{1}{N^q}\sum_{j=1}^{N^q}f_{\theta^{B}}(\mathbf{x}_{i,j}^q)$.
This task representation is then used as an additional input to the sentence classifier, which is trained end-to-end.
The meta-training process can be formally defined as:
\begin{equation}
    \theta^B_\star, \theta^c_\star \leftarrow \min_{\theta^B, \theta^c}\frac{1}{n}\sum_{i} \mathcal{L}\left(f_{\theta^B,\theta^c};\mathcal{D}_i^q, \mathbf{c}_i\right)
\end{equation}
Note that to enable unsupervised adaptation, ARM-CML learns to compute accurate task embeddings $\mathbf{c}_i$ from unlabeled data instead of using finetuning.
\section{Approach}
In this section, we present the proposed KGML framework (Fig.~\ref{fig:framework}), which allows us to enhance supervised and unsupevised adaptation methods described in the previous section with external knowledge extracted from a shared KB and.
In the following subsections, we elaborate the key components of KGML:
(1) extraction and representation of sentence-specific knowledge graphs~(KGs) and (2) knowledge fusion.

\subsection{KG Extraction and Representation}

For each sentence $\mathbf{x}_{i,j}$, we propose to extract a KG, denoted $\mathcal{G}_{i,j} = \{\mathcal{N}_{i,j}, \mathcal{E}_{i,j}\}$.
The nodes $\mathcal{N}_{i,j}$ of the graph correspond to entities in the corresponding sentence $\mathbf{x}_{i,j}$ and the edges $\mathcal{E}_{i,j}$ correspond to relations between these entities.
The relations between the entities are extracted from the KB shared across all tasks.
Notice that some entities are not directly related to each other in the KB. To enhance the density of graphs, we further ``densify'' the extracted KG with additional edges by constructing a k-nearest neighbor graph (k-NNG) based on the node embeddings.
More details of KG construction algorithm are provided in Appendix~\ref{sec:app_alg}.

To compute representations of the sentence-specific KGs, we use graph neural networks (GNN) \cite{kipf2016semi, GNN-survey}. In particular, we use GraphSAGE~\cite{GraphSage} as the forward propagation algorithm, which is formulated as follows:
\begin{equation}
\begin{aligned}
\mathbf{h}_{v}^{k} &=\sigma\left(\mathbf{W}_{1}^{k}\cdot \mathbf{h}_{v}^{k-1} +\mathbf{W}^{k}_{2} \cdot \mathbf{h}_{\mathcal{N}(v)}^{k}\right)\\
\text{s.t.}\; \mathbf{h}_{\mathcal{N}(v)}^{k} &=\textsc{Agg}_{k}\left(\left\{\mathbf{h}_{u}^{k-1}, \forall u \in \mathcal{N}(v)\right\}\right)
\end{aligned}
\end{equation}
where $\mathbf{W}^k (\forall k\in\{1,...,K\})$ are the weight matrices of the GNN, $\mathcal{N}(v)$ represents neighborhood set of node $v$ and $\mathbf{h}_{u}^{k}$ denotes the node representation in the k-th convolutional layer ($\mathbf{h}_v^0$ as the input feature). $\sigma$ and $\textsc{Agg}_k$ are functions of non-linearity and aggregator, respectively.

After passing each graph $\mathcal{G}_{i,j}$ into the graph neural network, we aggregate all node representations $\{\mathbf{h}_{v}^{K}\mid v\in \mathcal{N}_{i,j}\}$ and output the graph embedding $\mathbf{g}_{i,j}$ as the holistic representation of the knowledge graph.

\begin{algorithm}[h]
    \caption{KGML for Supervised Adaptation}
    \label{alg:kgml_l}
    \begin{algorithmic}[1]
    \REQUIRE Task distribution $p(\mathcal{T})$; Stepsize $\alpha$, $\beta$; Knowledge Base
    \STATE Randomly initialize parameter $\theta_0$, $\phi$
    \WHILE{not converge}
    \STATE Sample tasks \begin{small}$\{\mathcal{T}_i\}_{i=1}^{|I|}$\end{small}
    \FORALL{\begin{small}$\mathcal{T}_i$\end{small}}
    \STATE Sample support set \begin{small}$\mathcal{D}^{s}_i$\end{small} and query set  \begin{small}$\mathcal{D}^{q}_i$\end{small}
    \STATE Learn the sentence embeddings \begin{small}$f_{\theta^{B}}(\mathbf{x}_{i,j}^{s(q)})$\end{small}
    \STATE Extract the knowledge graph \begin{small}$\mathcal{G}_{i,j}^{s(q)}$\end{small} for each sentence \begin{small}$\mathbf{x}_{i,j}^{s(q)}$\end{small}
    \STATE For each graph, using GNN to learn the graph embedding \begin{small}$\mathbf{g}_{i,j}^{s(q)}$\end{small} via Eqn. (3)
    \STATE Fuse the sentence and graph embeddings via Eqn. (4) and obtain \begin{small}$\{\Tilde{f}_{\theta^B}(\mathbf{x}^{s(q)}_{i,j})\}_{j=1}^{N_{s(q)}}$\end{small}
    \STATE Compute the task specific parameter \begin{small}$\theta^c_i$\end{small} for MAML or compute the prototypes \begin{small}$\{\mathbf{c}_i^k\}_{k=1}^K$\end{small} for ProtoNet 
    \STATE Compute loss \begin{small}$\mathcal{L}(f_{\theta^c_i}(\{\Tilde{f}_{\theta^B}(\mathbf{x}^q_{i,j})\}_{j=1}^{N_q}),\mathbf{Y}_i^q)$\end{small}
    \ENDFOR
    \STATE Update all parameters \begin{small}$\theta^c, \theta^{B}, \phi:= \arg\min_{\theta_0, \theta_{B}, \phi}\frac{1}{|I|}\sum_{i}\mathcal{L}(f_{\theta^c_i}(\{\Tilde{f}_{\theta^B}(\mathbf{x}^q_{i,j})\}_{j=1}^{N_q}),\mathbf{Y}_i^q)$\end{small}
    \ENDWHILE
    \end{algorithmic}
\end{algorithm}

\subsection{Knowledge Fusion}
To bridge the distribution gap between meta-training and meta-testing stages, we integrate the information extracted from knowledge graph into the meta-learning framework. Assume the sentence representation is $f_{\theta^B}(\mathbf{x}_{i,j})$. For each sentence, we are motivated to design another aggregator $\mathrm{AGG}_{kf}$ to aggregate the information captured from the representation of sentence $f_{\theta^B}(\mathbf{x}_{i,j})$ and its corresponding knowledge graph representation $\mathbf{g}_{i,j}$. Specifically, 
the aggregator is formulated as: 
\begin{equation}
\label{eq:agg}
    \Tilde{f}_{\theta^B}(\mathbf{x}_{i,j})=\mathrm{AGG}_{kf}(f_{\theta^B}(\mathbf{x}_{i,j}), \mathbf{g}_{i,j})
\end{equation}
There are various selections of aggregators (e.g., fully connected layers, recurrent neural network), and we will detail the selection of aggregators in the Appendix D. Then, we replace the sentence representation $f_{\theta^B}(\mathbf{x}_{i,j})$ by $\Tilde{f}_{\theta^B}(\mathbf{x}_{i,j})$ in the meta-learning framework. We denote all parameters related to knowledge graph extraction and knowledge fusion as $\phi$. Notice that $\phi$ are globally shared across all task in MAML since we are suppose to connect the knowledge among them.
In Alg.~\ref{alg:kgml_l} and Alg.~\ref{alg:kgml_ul} (Appendix~\ref{sec:app_algall}), we show the meta-learning procedure of the proposed model under the settings of supervised and unsupervised adaption, respectively. 
\section{Experiments}
\begin{table*}[h]
\small
\caption{Performance for supervised and unsupervised adaptation methods. We report the averaged accuracy over 600 tasks (supervised adaptation)/all meta-testing users (unsupervised adaptation).}
\label{tab:label_adaptation}
\begin{center}
\begin{tabular}{c|cc|cc||c|cc}
\toprule
\multicolumn{5}{c||}{Supervised Adaptation} & \multicolumn{3}{c}{Unsupervised Adaptation}\\
Data & \multicolumn{2}{c|}{Amazon Review} & \multicolumn{2}{c||}{Huffpost} & Data & \multicolumn{2}{c}{Twitter}\\
Shot & 1-shot & 5-shot & 1-shot & 5-shot & User Ratio & 0.6 & 1.0\\\midrule
MAML & 44.35\% & 56.94\% & 39.95\% & 51.74\% & ERM & 62.91\% & 66.05\% \\
ProtoNet & 55.32\% & 73.30\%  & 41.72\% & 57.53\% & UW & 63.51\% & 64.13\%  \\
InductNet & 45.35\% & 56.73\% & 41.35\% & 55.96\% & ARM & 60.42\% & 60.42\%  \\
MatchingNet & 51.16\% & 69.89\% & 41.18\% & 54.41\%  & DRNN & 63.02\% & 64.02\% \\
REGRAB & 55.07\% & 72.53\% & 42.17\% & 57.66\% & - & - & - \\\midrule
\textbf{KGML-MAML} & 51.44\% & 58.81\% & \textbf{44.29\%} & 54.16\% & \textbf{KGML} & \textbf{64.92\%} & \textbf{67.00\%}\\
\textbf{KGML-ProtoNet} & \textbf{58.62\%} & \textbf{74.55\%} & 42.37\% & \textbf{58.75\%} & - & - & -\\
\bottomrule
\end{tabular}
\end{center}
\vspace{-1em}
\end{table*}
In this section, we show the effectiveness of our proposed KGML on three datasets and conduct related analytic study.
\subsection{Dataset Description}
Under the supervised adaptation, we leverage two text classification datasets. The first one is Amazon Review~\cite{ni2019justifying}, aiming to classify the category of each review. The second one is a headline category classification dataset -- Huffpost~\cite{huffpost}, aiming to classify the headlines of News. We apply the traditional N-way K-shot few-shot learning setting~\cite{finn2017model} on these datasets (N=5 in both Huffpost and Amazon Review). 

As for the unsupervised adaptation, similar to the settings in~\cite{zhang2020adaptive}, we use a federated sentiment classification dataset -- Twitter~\cite{caldas2018leaf}, to evaluate the performance of KGML. Each tasks in Twitter represents the sentences of one user. Detailed data descriptions are shown in Appendix~\ref{sec:app_data}.

\subsection{Experimental Settings}
For supervised adaptation, we compare KGML on five recent meta-learning algorithms, including MAML~\cite{finn2017model}, ProtoNet~\cite{snell2017prototypical}, Matching Network~\cite{vinyals2016matching} (MatchingNet), REGRAB~\cite{qu2020few}, Induction Network (InductNet)~\cite{geng2019induction}. We conduct the experiments under 1-shot and 5-shot settings and report the results of KGML with gradient-based meta-learning (KGML-MAML) and metric-based meta-learning (KGML-ProtoNet) algorithms.

Under the unsupervised adaptation scenario, KGML is compared with the following four baselines: empirical risk minimization (ERM), upweighting (UW), domain adversarial neural network (DANN)~\cite{ganin2015unsupervised}, and  adaptive risk minimization (ARM)~\cite{zhang2020adaptive}. Here, we report the performance with full users and 60\% users for meta-training.

On both scenarios, accuracy is used as the evaluation metric and all baselines use  ALBERT~\cite{lan2019albert} as encoder. WordNet~\cite{miller1995wordnet} is used as the knowledge graph. All other hyperparameters are reported in Appendix~\ref{sec:app_hyper}.

\subsection{Overall Performance}
The overall performance of all baselines and KGML are reported in Table~\ref{tab:label_adaptation}. The results indicate that KGML achieves the best performance in all scenarios by using knowledge bases to bridge the gap between the meta-training and meta-testing tasks. Additionally, under the supervised adaptation scenario, the improvements of Amazon Review are larger than that in Huffpost under the 1-shot setting, indicating that the former has a larger gap between meta-training and meta-testing tasks. One potential reason is that the number of entities of Amazon review is more than Huffpost headlines, resulting in more comprehensive knowledge graphs. Another interesting finding is that ARM hurts the performance under the unsupervised adaptation. However, with the help of the knowledge graph, KGML achieves the best performance, corroborating its effectiveness in learning more transferable representations and further enabling efficient unsupervised adaptation.

\subsection{Ablation Study}
We conduct ablation studies to investigate the contribution of each component in KGML. Two ablation models are proposed: I. replacing the aggregator $\mathrm{AGG}_{kf}$ with a simple feature concatenator; II. removing extra edges in KG, which are introduced by k-nearest neighbor graph. The performance of each ablation model and the KGML of Amazon and Huffpost are reported in Table~\ref{tab:ablation}. We observe that (1) KGML outperforms model I, demonstrating the effectiveness of the designed aggregator; (2) Comparing between KGML with model II, the results show that KNN boosts performance. One potential reason is that KNN densifies the whole network according to the entities' semantic embeddings learned from the original WordNet, which explicitly enriches the semantic information of the neighbor set of each entity. It further benefits the representation learning process and improves the performance.
\begin{table}[h]
\small
\caption{Ablation study (1-shot scenario). Backbone: base meta-learning algorithm}
\vspace{-0.5em}
\label{tab:ablation}
\begin{center}
\setlength{\tabcolsep}{1.7mm}{
\begin{tabular}{l|l|cc}
\toprule
Ablations & Backbone & Amazon & Huffpost\\\midrule
\multirow{2}{*}{I. Remove $\mathrm{AGG}_{kf}$} & MAML & 45.68\% & 41.55\% \\
& ProtoNet & 57.94\% & 41.71\% \\\midrule
\multirow{2}{*}{II. Remove KNN} & MAML & 51.07\% & 41.20\%\\
& ProtoNet & 57.80\% & 41.91\%  \\\midrule
KGML & MAML & 51.44\% & \textbf{44.29\%} \\
KGML & ProtoNet & \textbf{58.62\%} & 42.37\%\\
\bottomrule
\end{tabular}}
\end{center}
\vspace{-1em}
\end{table}

\subsection{Robustness Analysis}
\begin{figure}[h]
\centering
	\begin{subfigure}[c]{0.23\textwidth}
		\centering
		\includegraphics[height=28mm]{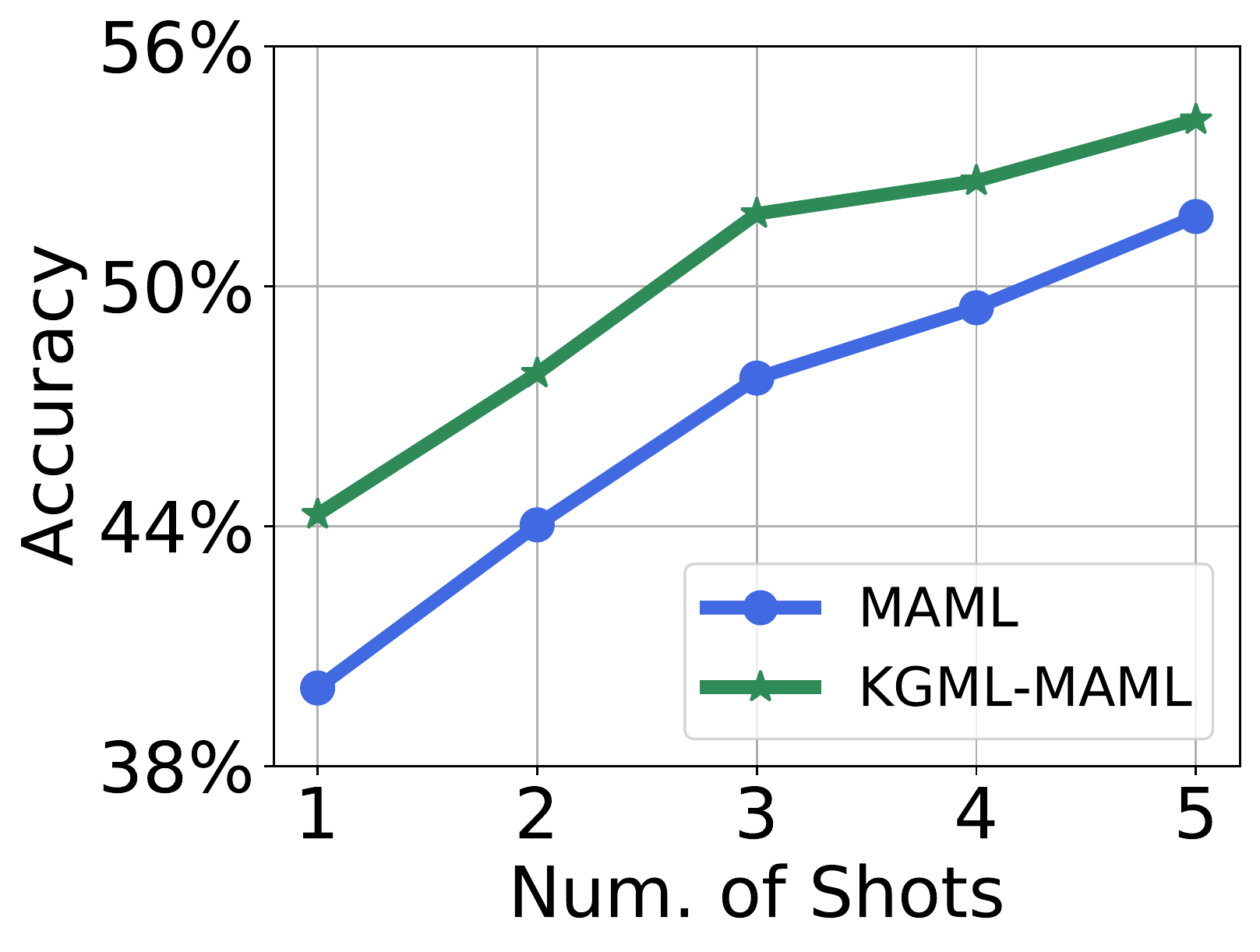}
		\caption{Huffpost\label{fig:huffpost}}
	\end{subfigure}
	\begin{subfigure}[c]{0.23\textwidth}
		\centering
		\includegraphics[height=28mm]{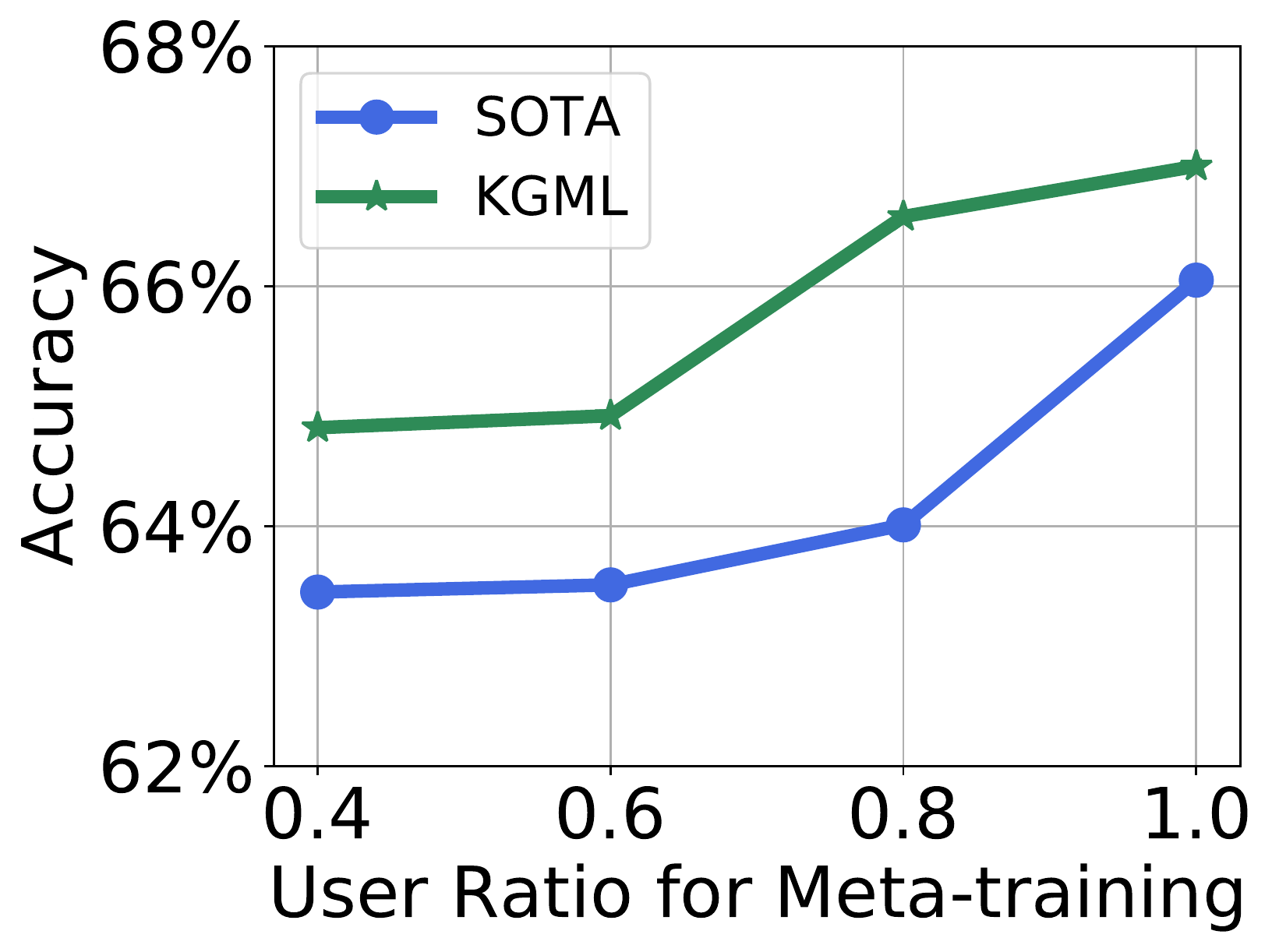}
		\caption{Twitter\label{fig:twitter}}
	\end{subfigure}
	\vspace{-0.3em}
	\caption{Robustness analysis. SOTA: best baseline}
	\vspace{-0.5em}
\end{figure}

In this subsection, we analyze the robustness of KGML under different settings. Specifically, under supervised adaptation, we change the number of shots in Huffpost. Under unsupervised adaptation, we reduce the number of training users in Twitter. The performance are illustrated in Figure~\ref{fig:huffpost} and Figure~\ref{fig:twitter}, respectively (see the comparison between Huffpost-ProtoNet and ProtoNet in Appendix~\ref{sec:app_figure}). From these figures, we observe that KGML consistently improves the performance in all settings, verifying its effectiveness to improve the generalization ability.

\subsection{Discussion of Computational Complexity}
We further conduct the analysis of computational complexity and reported the meta-training time per task in Table~\ref{tab:time}, where the results of supervised adaptation are performed under the setting of Huffpost 5-shot.
Though KGML increases the meta-training time to some extent, the who training process can be finished within 1-2 hours. Thus, the additional computational cost seems to be a reasonable trade-off for accuracy.
\begin{table}[h]
\small
\caption{Results of meta-training time per task.}
\vspace{-1em}
\label{tab:time}
\begin{center}
\setlength{\tabcolsep}{1.7mm}{
\begin{tabular}{l|c|c}
\toprule
Model & Supervised (MAML) & Unsupervised\\\midrule
w/o KG & 0.297s & 0.146s
\\
with KG & 0.407s & 0.181s \\
\bottomrule
\end{tabular}}
\end{center}
\vspace{-1.5em}
\end{table}
\section{Conclusion}
In this paper, we investigated the problem of meta-learning on low-resource text classification, and propose a new method KGML. Specifically, by learning the representation from extracted sentence-specific knowledge graphs, KGML bridges the gap between meta-training and meta-testing tasks, which further improves the generalization ability of meta-learning. KGML is compatible with supervised and unsupervised adaptation and the empirical experiments on three datasets demonstrate its effectiveness over state-of-the-art methods.
\section*{Acknowledgments}
This work is partially supported by NSF awards IIS-\#1617583, IIS-\#1955532, CNS-\#2008248 and NGA HM-\#04762010002. The views and conclusions contained in this paper are those of the authors and should not be interpreted as representing any funding agencies.
\bibliography{ref}
\bibliographystyle{acl_natbib}
\appendix
\section{Detailed Descriptions of Sentence-specific KG Construction}
\label{sec:app_alg}
To construct a holistic knowledge graph with all the entities and relations, we first use the existing knowledge graph (i.e., WordNet~\cite{miller1995wordnet}) as the sparse knowledge base $\mathcal{G}^{base}$. Some entities may be the nodes with few interactions or even isolated. Thus, to connect all entities, the node embeddings in the knowledge base are then used to construct a K-NN graph $\mathcal{G}^{knn}$, which is further combined with the base knowledge graph, rendering the dense knowledge graph $\mathcal{G} = \mathcal{G}^{knn}\cup \mathcal{G}^{base}$.

For each sentence $\mathbf{x}_{i,j}$, we use its entities to query the knowledge graph $\mathcal{G}$, which returns the entity embeddings $\mathcal{N}_{i,j}$ and a adjacency matrix $\mathcal{A}_{i,j}$. Each element in $\mathcal{A}_{i,j}$ represents the shortest distance of the corresponding entities. Inspired by Occam's Razor criterion, we compute the Minimum Spanning Tree (MST)~\cite{wiki:Minimum_spanning_tree} w.r.t all the target entities (other entities and relations in the chosen path are included) as the concise and informative graphical representation of the sentence. 
In Alg.~\ref{alg:kg_extract}, we illustrate the whole process of the knowledge graph.
\begin{algorithm}[h]
    \caption{Knowledge Graph Extraction}
    \label{alg:kg_extract}
    \begin{algorithmic}[1]
    \REQUIRE Dense knowledge graph $\mathcal{G}$
    \FOR{each sentence $\mathbf{x}_{i,j}$}
    \STATE Use the entities $\mathbf{x}_{i,j}$ to query $\mathcal{G}$ and obtain entity embeddings $\mathcal{N}_{i,j}$ and adjacency matrix $\mathcal{A}_{i,j}$
    \STATE Apply MST algorithm on $\mathcal{A}_{i,j}$, which returns $\mathrm{T}$, the minimum spanning tree w.r.t the entities in $\mathbf{x}_{i,j}$.
    \STATE Construct the knowledge graph $\mathcal{G}_{i,j}$ by including the selected nodes and edges on the path of $\mathrm{T}$, i.e., $\mathcal{G}_{i,j} = \{(r, s)\mid\exists (u,v) \in \mathrm{T}, (r, s) \in \operatorname{ShortestPath}(u, v)\}$.
    \ENDFOR
    \end{algorithmic}
\end{algorithm}
\section{Pseudocodes of KGML}
\label{sec:app_algall}
In this section, we add the pseudocode for unsupervised adaptation in Alg.~\ref{alg:kgml_ul}.

\begin{algorithm}[h]
    \caption{KGML for Unsupervised Adaptation}
    \label{alg:kgml_ul}
    \begin{algorithmic}[1]
    \REQUIRE Task distribution $p(\mathcal{T})$; Stepsize $\beta$; Knowledge Base
    \STATE Randomly initialize parameter $\theta_0$, $\phi$
    \WHILE{not converge}
    \STATE Sample tasks \begin{small}$\{\mathcal{T}_i\}_{i=1}^{|I|}$\end{small}
    \FORALL{\begin{small}$\mathcal{T}_i$\end{small}}
    \STATE Sample query set  \begin{small}$\mathcal{D}^{q}_i$\end{small} from the task \begin{small}$\mathcal{T}_i$\end{small}
    \STATE Learn the sentence embeddings \begin{small}$f_{\theta^{B}}(\mathbf{x}_{i,j}^q)$\end{small}
    \STATE Extract the knowledge graph \begin{small}$\mathcal{G}^q_{i,j}$\end{small}
    \STATE For each graph, using GNN to learn the graph embedding \begin{small}$\mathbf{g}^q_{i,j}$\end{small} via Eqn. (3)
    \STATE Fuse the sentence and graph embeddings and obtain the final embedding \begin{small}$\{f_{\theta^{B}}(\mathbf{x}_{i,j}^q)\}_{j=1}^{N_s}$\end{small}
    \STATE Calculate the contextual vector $\mathbf{c}_i$ and compute loss \begin{small}$\mathcal{L}(f_{\theta_0}(\{f_{\theta^{B}}(\mathbf{x}_{i,j}^q)\}_{j=1}^{N_q}, \mathbf{c}_i),\mathbf{Y}_i^q)$\end{small}
    \ENDFOR
    \STATE Update all parameters \begin{small}$\theta^c, \theta^{B}, \phi:= \arg\min\frac{1}{|I|}\sum_{i}\mathcal{L}(f_{\theta^c}(\{\Tilde{f}_{\theta^{B}}(\mathbf{x}_{i,j}^q)\}_{j=1}^{N_q}, \mathbf{c}_i),\mathbf{Y}_i^q)$\end{small}
    \ENDWHILE
    \end{algorithmic}
\end{algorithm}

\section{Data Statistics}
\label{sec:app_data}
For supervised adaptation, we use Amazon Review and Huffpost to evaluate the performance. Amazon Review contains 28 classes, and the number of classes for meta-training, meta-validation, and meta-testing are 15, 5, 8, respectively. The Huffpost dataset includes 41 classes in total, and we use 25, 6, 10 classes for meta-training, meta-validation, and meta-testing, respectively. In terms of the unlabeled adaptation, the number of Twitter users for meta-training, meta-validation, and meta-testing are 741, 92, 94, respectively.
\section{Hyperparameter Settings}
\label{sec:app_hyper}
For all the supervised adaptation and the unlabeled adaption experiments, we use ALBERT~\cite{lan2019albert} as the sentence encoders and GraphSAGE~\cite{GraphSage} as the graph encoders. All hyperparameters are selected via the performance on the validation set.
\subsection{Supervised Adaptation}
The GNN used contains two layers, where the number of neurons is 64 and 16, respectively. We adopt two fully connected layers with ReLU as activation layer for the adaptation layers, where the number of neurons is 64 for each layer. The aggregator $\mathrm{AGG}_{kf}$ is designed as the one fully connected layer. We set the inner-loop learning rate $\alpha$ and outer-loop learning rate $\beta$ as 0.01 and 2e-5, respectively. The number of steps in the inner loop is set as 5. We use Adam~\cite{kingma2014adam} for outer loop optimization. The maximum number of epochs for huffpost and Amazon Review is 10,000 and 4,000, respectively.
\subsection{Unsupervised Adaptation}
For the sentence encoder, the number of output dimensions is set as 240. 
The GNN is composed of two convolution layers, where each layer contains 64 neurons, and $\mathrm{AGG}_{k}$ is designed as a mean pool operation. 
We use one fully connected layer for the final aggregation $\mathrm{AGG}_{kf}$. 
In the training phase, the learning rate is set $\beta$ as 1e-4, and we use Adam~\cite{kingma2014adam} optimizer with weight decay 1e-5. 
The contextual support size and meta batch size are 50 and 2, respectively. 
\section{Additional Results of Robustness Analysis}
\label{sec:app_figure}
\begin{figure}[h]
\centering
	\includegraphics[height=35mm]{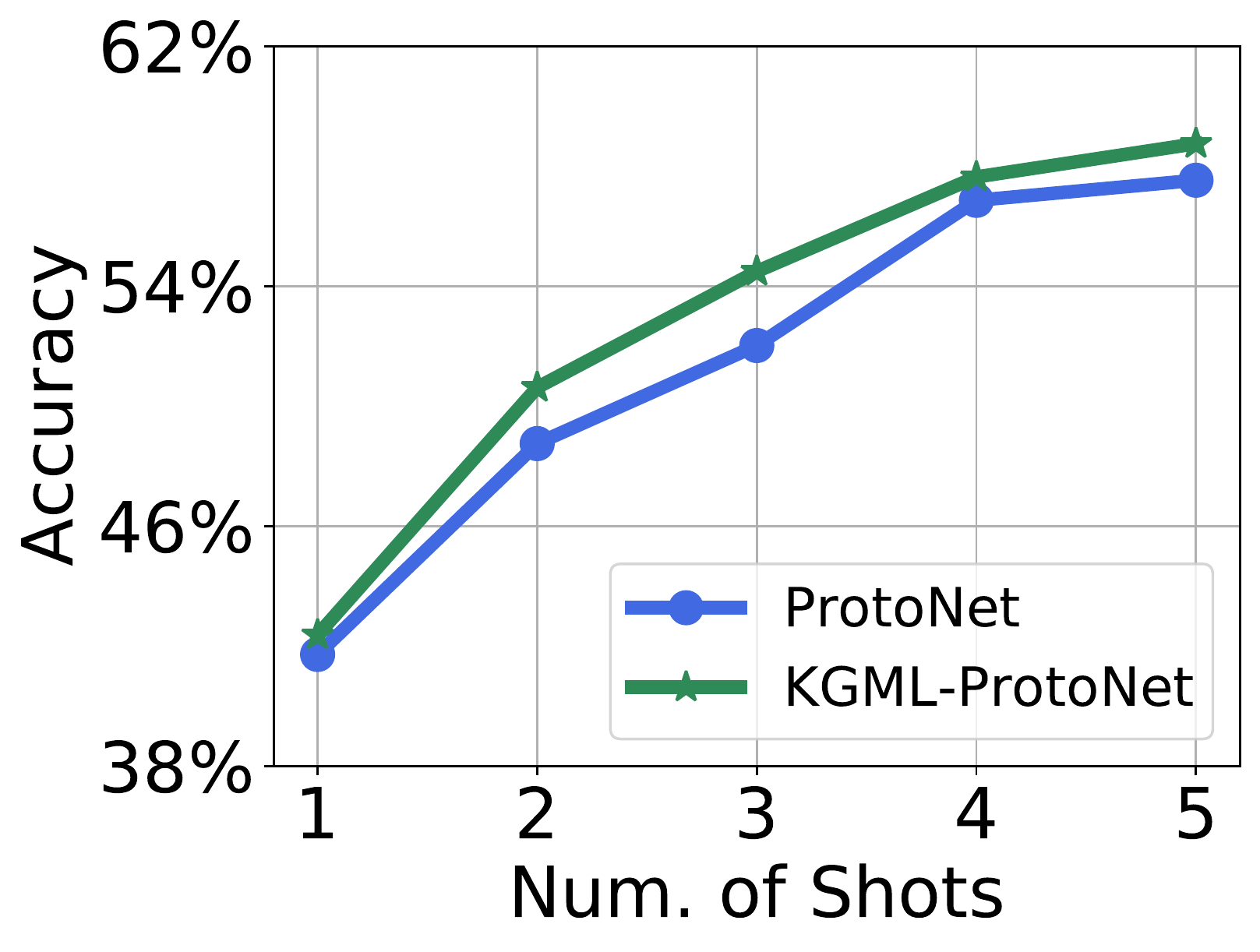}
	\caption{Additional Robustness analysis. }
	\label{app:fig_protonet}

\end{figure}

In Figure~\ref{app:fig_protonet}, we show the comparison between Huffpost-ProtoNet and ProtoNet w.r.t. the number of shots. The results further demonstrate the effectiveness of KGML.
\end{document}